%% file: 0_main_camera_ready.tex
\documentclass[sigconf,nonacm=true,urlbreakonhyphens=true,balance=true]{acmart}
\AtBeginDocument{%
  \providecommand\BibTeX{{%
    \normalfont B\kern-0.5em{\scshape i\kern-0.25em b}\kern-0.8em\TeX}}}

\copyrightyear{2021}
\acmYear{2021}
\setcopyright{iw3c2w3}
\acmConference[WWW '21]{Proceedings of the Web Conference 2021}{April 19--23, 2021}{Ljubljana, Slovenia}
\acmBooktitle{Proceedings of the Web Conference 2021 (WWW '21), April 19--23, 2021, Ljubljana, Slovenia}
\acmPrice{}
\acmDOI{10.1145/3442381.3449809}
\acmISBN{978-1-4503-8312-7/21/04}

\usepackage{amsbsy}
\usepackage{soul}
\usepackage{comment}
\usepackage{hhline}
\usepackage{multirow}
\usepackage{enumitem}
\usepackage[belowskip=2pt,aboveskip=2pt]{caption}
\usepackage[ruled,vlined,linesnumbered]{algorithm2e}
\SetAlFnt{\small}
\SetAlCapFnt{\small}
\SetAlCapNameFnt{\small}
\usepackage{algorithmic}
\algsetup{linenosize=\tiny}
 
\begin{document}

\title{One Detector to Rule Them All}
\subtitle{Towards a General Deepfake Attack Detection Framework}
\author{Shahroz Tariq}
\email{shahroz@g.skku.edu}
\orcid{1234-5678-9012}
\affiliation{%
  \institution{Sungkyunkwan University}
  \city{Suwon}
  \state{South Korea}
}

\author{Sangyup Lee}
\email{sangyup.lee@g.skku.edu}
\affiliation{%
  \institution{Sungkyunkwan University}
  \city{Suwon}
  \state{South Korea}
}

\author{Simon S. Woo}
\email{swoo@g.skku.edu}
\affiliation{%
  \institution{Sungkyunkwan University}
  \city{Suwon}
  \state{South Korea}
}


\begin{abstract}
  \input{1_abstract}
\end{abstract}




\maketitle
\input{2_introduction}

\input{3_0_related}
\input{3_1_AttackModel}

\input{4_approach}
\input{5_0_Experiment}

\input{5_results}

\balance
\input{6_conclusion}

\begin{acks}
We thank the reviewers for the insightful reviews and Siho Han for proofreading.
This work is supported by Institute of Information \& communications Technology Planning \& Evaluation (IITP) grant funded by Korea government (MSIT) (No.2021-0-00017) and (No.2019-0-01343, Regional strategic industry convergence security core talent training business),
the High-Potential Individuals Global Training Program (No.2019-0-01579) supervised by IITP,
and the Basic Science Research Program through National Research Foundation of Korea grant funded by MSIT (No.2020R1C1C1006004).
\end{acks}
\bibliographystyle{ACM-Reference-Format}
\bibliography{references.bib}
\end{document}

%% file: 1_abstract.tex
Deep learning-based video manipulation methods have become widely accessible to the masses. With little to no effort, people can quickly learn how to generate deepfake (DF) videos. While deep learning-based detection methods have been proposed to identify specific types of DFs, their performance suffers for other types of deepfake methods, including real-world deepfakes, on which they are not sufficiently trained. In other words, most of the proposed deep learning-based detection methods lack transferability and generalizability. Beyond detecting a single type of DF from benchmark deepfake datasets, we focus on developing a generalized approach to detect multiple types of DFs, including deepfakes from unknown generation methods such as DeepFake-in-the-Wild (DFW) videos. To better cope with unknown and unseen deepfakes, we introduce a Convolutional LSTM-based Residual Network (CLRNet), which adopts a unique model training strategy and explores spatial as well as the temporal information in a deepfakes. Through extensive experiments, we show that existing defense methods are not ready for real-world deployment. Whereas our defense method (CLRNet) achieves far better generalization when detecting various benchmark deepfake methods (97.57\% on average). Furthermore, we evaluate our approach with a high-quality DeepFake-in-the-Wild dataset, collected from the Internet containing numerous videos and having more than 150,000 frames. Our CLRNet model demonstrated that it generalizes well against high-quality DFW videos by achieving 93.86\% detection accuracy, outperforming existing state-of-the-art defense methods by a considerable margin.

%% file: 2_introduction.tex
\begin{figure}[t!]
    \centering
    \includegraphics[width=1\linewidth]{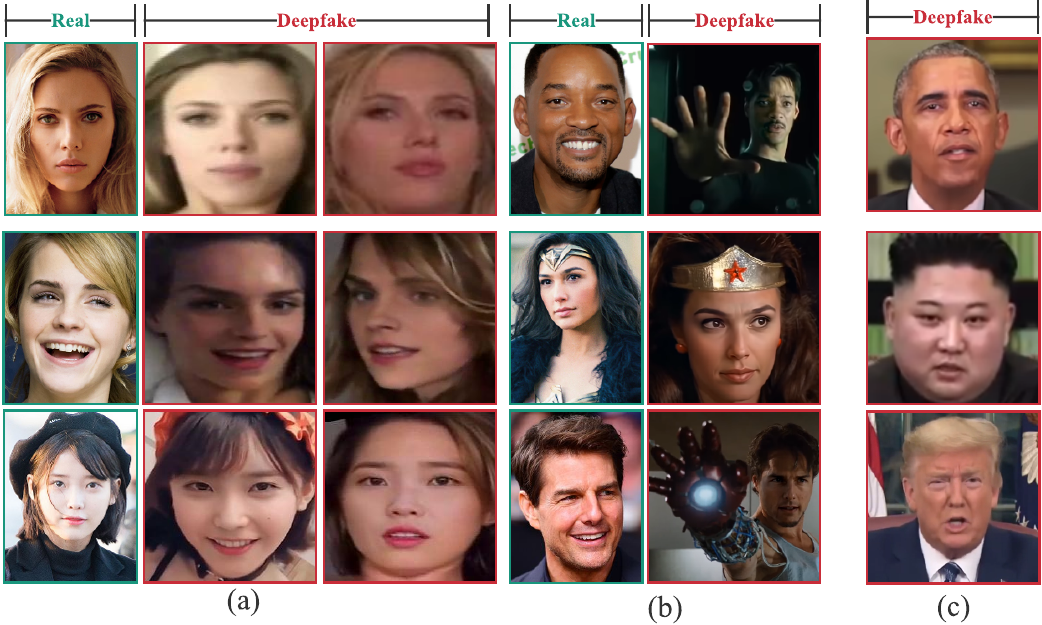}
    \caption{Illustration of the DeepFake-in-the-Wild (DFW).
    The deepfakes shown in (a) are of the celebrities created by various unknown generation methods.
    In (b) famous movie characters are replaced with some other celebrity and (c) contains deepfakes of famous political figures.}
    \label{fig:DFW}
\end{figure}

\section{Introduction}
\label{sec:intro}
Deep learning-based methods for synthetic or manipulated video generation have arisen tremendously in the last few years. These recent methods can generate photorealistic images that can easily deceive the average human eyes~\cite{FaceForensics++,Face2Face,FaceSwap,Deepfakes,NeuralTextures,PGGAN}. Due to their ability, these methods have many applications in computer vision or graphics-related disciplines, such as human face generation~\cite{PGGAN} and photorealistic scenery generation~\cite{GauGAN}. However, this innovation is also susceptible to potential abuse; many people with malicious intentions have taken advantage of these methods to generate fake videos of female celebrities and the general public~\cite{news1,news2,news3,news4} through various approaches~\cite{Face2Face,FaceSwap,NeuralTextures,Deepfakes}. This has started causing major social issues: a recent study claimed that 96\% of the deepfakes originate from porn videos~\cite{news5}. They come under the same umbrella of so-called Deepfakes. However, detecting these deepfakes or forged images/videos is challenging due to the lack of data. Also, designing a generalized classifier that performs well universally on different types of deepfakes is what we desperately need today~\cite{ShahrozAWS}.

\begin{figure*}[t!]
    \centering
    \includegraphics[width=0.9\linewidth]{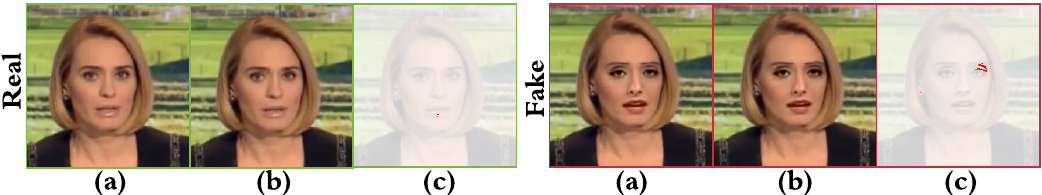}
    \caption{Difference between consecutive frames of a deepfake video. (a) and (b) are the $n^{\text{th}}$ and $(n+1)^{\text{th}}$ frames. For real videos, the difference between frames is small, while for deepfake videos, it is bigger (c), as highlighted in red.}
    \label{fig:inconsistency}
\end{figure*}

As an effort to increase the amount of available data, the research community has been recently releasing numerous deepfake datasets to assist other researchers in developing detection mechanisms for these deepfakes. The most pioneering work is the FaceForensics++ dataset~\cite{FaceForensics++} partly developed by Google. The FaceForensics++~\cite{FaceForensics++} dataset contained real and five deepfake datasets.
In 2019, Facebook launched a deepfake detection challenge with prize money of one million U.S. dollars to accelerate research in this field~\cite{DFDCWebsite}. Recently, Li et al.~\cite{CelebDFDataset} released the CelebDF dataset, which contains 5,639 deepfake videos.
Although, these benchmark datasets help in improving the performance and diversifying methods. 
However, it would be impractical to produce a dataset of considerable size for every novel deepfake generation method to enable the training of deep neural nets. While several deepfake detection methods achieve high test accuracy on a single deepfake dataset~\cite{ForensicTransfer,DFD1,DFD3,Shahroz2}, they have shown low detection accuracy on new deepfake methods developed for malicious purposes that were not introduced during the training phase (see Section~\ref{sec:ood_performance}). While several deepfake detection methods achieve high test accuracy on a single deepfake dataset, they have shown low detection accuracy on new deepfake methods developed for malicious purposes that were not introduced during the training phase. Therefore, it is natural that such methods also perform poorly on real-world DeepFake-in-the-Wild (DFW; see Fig.~\ref{fig:DFW}) videos because some other deepfake methods may have developed them. There is no extensive research on evaluating the generalizability of the deepfake detection classifiers against DFW videos. Therefore, beyond detecting a single type of deepfake or deepfakes from benchmark datasets. Our work's primary focus is to investigate a training strategy and classifier that can detect deepfakes from multiple benchmark datasets (e.g., FaceSwap, DeepFake, and Face2Face) and unknown generation methods (DFW videos).

In order to address this low generalizability of deepfake detectors, weakly-supervised approaches for transfer learning and domain adaptation, as well as few-shot learning-based approaches~\cite{ForensicTransfer} have proposed. However, they have not achieved satisfactory performance due to the small number of training samples and catastrophic forgetting~\cite{catastrophic1}. Also, most of the deepfake detection methods~\cite{ForensicTransfer, FaceForensics++,Xception,Shahroz1,Shahroz2,khalid2020oc,jeon2020t,SAMGAN,Hyeonseong1} independently extract frames (images) from videos for training and testing, constituting a single frame-based detection method, but do not explore the temporal aspects in a frame sequence, which can also be extremely useful for the detection process. Following careful observation, we were able to discern with our naked eyes tiny artifacts within consecutive frames of deepfake videos, as shown in Fig.~\ref{fig:inconsistency}.

Even though spatial information is the most important for detecting deepfakes, as shown by previous research~\cite{afchar2018mesonet,Shahroz1,ForensicTransfer}, the temporal information can also help the model find deepfakes. Therefore, we hypothesize that using spatial and temporal information can result in better detection performance.

To research a generalized and practical approach against benchmarks and DFW, we propose a convolutional LSTM based Residual Network (CLRNet) and several training strategies. The convolutional LSTM cells are effective in handling spatiotemporal information~\cite{ConvLSTM1,tariq2019convlstm,ShahrozConvlstm1,ShahrozConvlstm2}.
CLRNet takes a sequence of consecutive images from a video as an input to learn from the spatial as well as temporal information that helps in detecting unnatural artifacts present within consecutive frames and within a frame of deepfake videos.
To improve generalizability in terms of detection performance regardless of the generation method used, we develop detection and defense strategies under three training conditions: 1) single domain learning, 2) merge learning, and 3) transfer learning. We investigate how these training strategies affect model generalizability and employ them to detect different types of deepfakes altogether. We compare the performance of our CLRNet with those of other state-of-the-art baselines and show that our method outperforms them with better generalizability, achieving a 97.57\% accuracy on average for all benchmark deepfake datasets. 

To evaluate the classifiers, we considered three types of attackers in our threat model. 1) In-domain attacker: Trained and tested with the same benchmark dataset, 2) Out-of-Domain attacker: Trained and tested with different benchmark datasets, and 3) Open-domain attacker: Trained with benchmark dataset and tested with DeepFake-in-the-Wild videos\footnote{\textbf{Training and Testing set:} \textit{The training and testing sets have no overlapping data samples, even if they belong to the same dataset.}}. The details of our threat model are given in Section~\ref{sec:attack}). We evaluated unseen deepfakes generated by unknown methods such as DFW using more than 150,000 frames from 200 high-quality deepfake videos on the Internet and social media websites after consulting Institutional Review Board (IRB)\footnote{\textbf{Note on DFW videos:} \textit{We have not stored any of those videos on our machines. We directly evaluated them from their respective website by reading them through our code.}}.

We systematically evaluate the effectiveness of existing defenses on the DFW videos dataset and observe that they are not ready for deployment in the real-world. Whereas, CLRNet successfully detects different deepfake in-the-wild videos, yielding a 93.22\% accuracy as compared to 80.22\% accuracy of the best baseline method MesoNet. This demonstrates the importance of 1) exploring spatial and temporal information in deepfake videos, and 2) selecting a viable defense training strategy.

Our code, dataset collection process, and additional results are available here\footnote{\textbf{Code: }\textit{\url{https://github.com/shahroztariq/CLRNet}}}. The main contributions of our work are  as follows:

\begin{itemize}[leftmargin=10pt]
    \item \textbf{Convolutional LSTM-based Residual Network.} We propose a novel architecture based on Convolutional LSTM cells and Residual blocks for deepfake detection, leveraging spatial as well as temporal information present in consecutive frames extracted from  deepfake videos. We demonstrate state-of-the-art detection accuracy (97.51\%) on benchmark deepfake datasets with CLRNet. 
    
    \item \textbf{Generalization using Training Strategies.} We propose an effective, multiple training strategy (Merge- vs. Transfer-learning) to detect different types of deepfakes and demonstrate the advantages of the proposed strategy through rigorous experimentation. 
    
    \item \textbf{DeepFake-in-the-Wild (DFW) Detection.} Beyond evaluating benchmark deepfake datasets. When evaluated more than 150,000 frames from 200 high quality real-world DFW non-pornographic videos from multiple online sources, our CLRNet outperforms baseline methods, achieving a 93.22\% accuracy. 

\end{itemize}

%% file: 3_0_related.tex
\section{Background and Related Work}
\label{sec:related}
\begin{figure*}[t]
    \centering
    \includegraphics[width=1\linewidth]{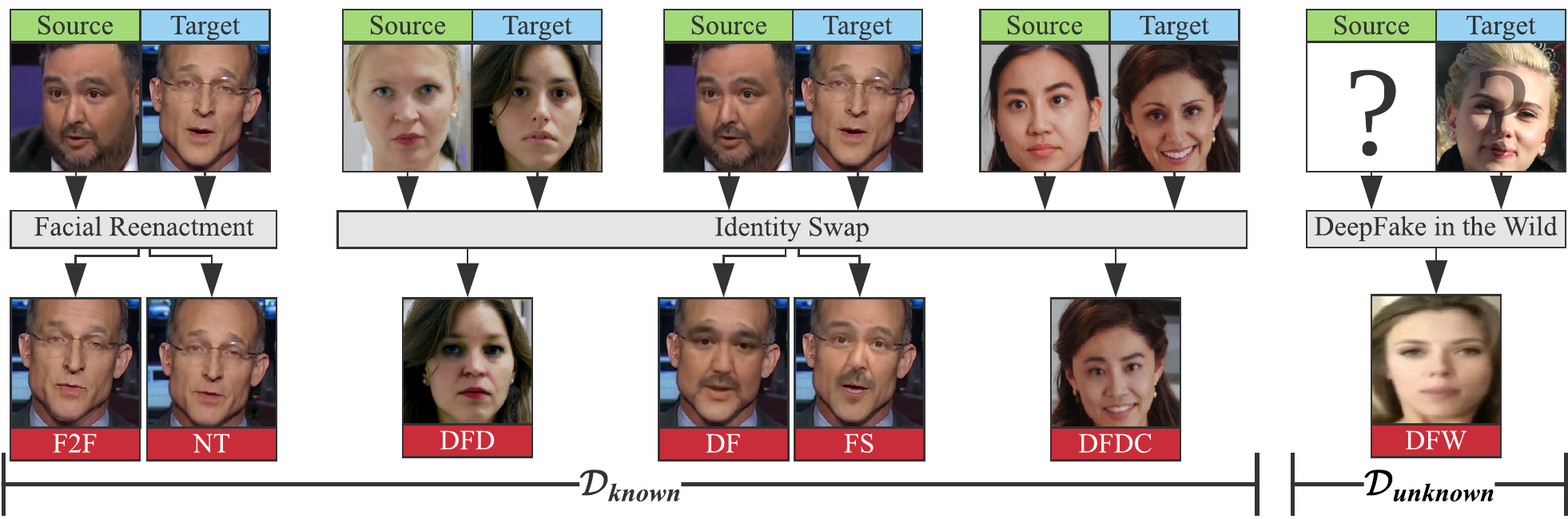}
    \caption{Different deepfake generation methods. Facial Reenactment and Identity Swap are general, high-level categories of deepfakes. The DeepFake-in-the-Wild contains deepfakes whose generation methods are unknown.}
    \label{fig:generation_types}
\end{figure*}

\subsection{Deepfake Generation Methods and Datasets}
Recently, there is a surge of deepfake or AI-generated synthetic videos in the online community~\cite{news1,news2,news3,news4,news5}.
Mirsky and Lee~\cite{Mirsky_WenkeLee_Survery} showed that there are various methods to generate these fake videos, but these generation methods are generally called deepfakes. Here, we would like to make some distinction between the different methods used to generate these deepfakes. We can divide the video generation methods into the following two major classes: 1) computer graphics-based methods such as Face2Face~\cite{Face2Face} and FaceSwap~\cite{FaceSwap}, and 2) Deep Learning-based methods such as Deepfakes~\cite{Deepfakes} and Neural Textures~\cite{NeuralTextures}. However, the DeepFake Detection Dataset~\cite{GoogleDFDdataset} by Google and Deepfake Detection Challenge dataset~\cite{DFDC} by Facebook cannot be assigned to any of the two classes, as they contain deepfakes from different identity swapping (faceswap) methods~\cite{DFDC,GoogleDFDdataset}. 
Furthermore, we evaluated with DeepFake-in-the-Wild videos, which contains deepfakes of numerous celebrities and politicians, generated by unknown methods, as shown in Fig.~\ref{fig:DFW}. We illustrate a high-level overview of these generation methods in Fig.~\ref{fig:generation_types}. Details of different deepfake datasets and their generation methods are discussed in the following sections.

\noindent
\textbf{DeepFake (DF).}
The word deepfakes is also name of a specific method to generate deepfake that has been distributed across online forums~\cite{Deepfakes}. To distinguish them, we refer to this dataset in our paper by DeepFake.
In the DeepFake method, the target face is replaced by a face that is observed in a source video or image series.

\noindent
\textbf{FaceSwap (FS).} 
It is a computer graphics-based method by which the face region is transferred from source to a target video.

\noindent
\textbf{Face2Face (F2F).}
It is a method of facial reenactment that transfers the expressions of a source video to a target video while preserving the target person's identity.

\noindent
\textbf{Neural Textures (NT).}
The source video data is used to grasp the target person's neural texture, including a rendering network. 
Only the facial expressions corresponding to the mouth region are modified, i.e., the eye region remains unchanged.

\noindent
\textbf{Deepfake Detection (DFD).}
Google/Jigsaw's Deepfake Detection (DFD) dataset consists of 363 real and 3,000 deepfakes of 28 paid actors. 
Every person has been given tasks, such as walking
while using various expressions such happy or angry.
A variety of off-the-shelf faceswap methods are applied to build this dataset~\cite{GoogleDFDdataset}.

\noindent
\textbf{Deepfake Detection Challenge (DFDC).}
It is one of the most extensive deepfake datasets comprised of 100,000 videos of different quality, views, lighting, and scenes. Similar to DFD, a variety of faceswap methods are used to develop this dataset~\cite{DFDC}.

\noindent
\textbf{DeepFake-in-the-Wild (DFW). } 
Currently, the majority of the deepfake videos on the internet are of women and mostly are of celebrities~\cite{news5}. These celebrity deepfake videos are typically abused for the production of pornography and other malicious intends. These deepfakes are readily available and easily accessible on numerous websites, including social network services. As shown in Fig.~\ref{fig:DFW}, we focus on the following three types of DFW videos, which are the most popularly used for creating deepfakes:First, deepfakes generated using celebrity faces pasted onto random online videos. Second, deepfakes with a movie character is replaced with another celebrity. Third, political speech video manipulated using deepfake methods. Most of the DFW videos available on online video hosting websites such as YouTube and Bilibili are of low quality. Therefore, these videos are easily detectable by most deepfake detectors and humans as fakes. Therefore, we focused on finding high-quality and polished DeepFake-in-the-Wild videos, which can fool both humans and machines alike.\footnote{\textbf{Note:} \textit{We only evaluated these DFW videos by directly reading frames from the online video source using our code. We have not stored them anywhere in our possession.}}.
Before evaluating these DFW videos, we first extensively consulted with the Institutional Review Board (IRB) in our institution for more in-depth discussion on ethics and privacy issues regarding DFW videos (see Section~\ref{sec:discussion}). 
Nothing concrete can be said about the generation methods used to create these deepfake videos, and it is nearly impossible to find the original source and target videos used to generate these deepfake videos, since we do not have ground truth (see DFW in Fig.~\ref{fig:generation_types}). However, based on our visual analysis and experimental results, we believe that a variety of unknown deepfake generation methods. Also, it is possible that the DFW dataset has been generated by a mix of different and/or novel methods (see Section~\ref{sec:res} for details).

\noindent
\textbf{Defining Known and Unknown Datasets. }
As shown in Fig.~\ref{fig:generation_types}, the deepfake generation methods are known, at the high-level, as Facial Reenactment and Identity Swap.
In \textit{Facial Reenactment} methods, such as F2F and NT, the source is used to drive the expression, gaze, mouth, and pose of the target. In \textit{Identity Swap} methods, such as DF, FS, DFD, and DFDC, some parts or the entire face of the target is swapped with the source face while preserving the identity of the target. \textit{DeepFake-in-the-Wild} videos are possibly generated using numerous off-the-shelf deepfake generation methods. Therefore, it is extremely difficult to find the source and target used for this generation method, further complicating its analysis and detection. The generation methods for DF, FS, F2F, NT, and DFD datasets are known at a high level, as shown with Facial Reenactment and Identity Swap in Fig.~\ref{fig:generation_types}. Let $\mathcal{D}_{known}$ be the set of datasets created using known deepfake generation methods as follows:
\begin{equation}
    \label{eq:known}
    \small
    \mathcal{D}_{known} = \left \{ DF, FS, F2F, NT, DFD \right\},
\end{equation}
Meanwhile, the generation method for the DeepFake-in-the-Wild dataset is unknown, as shown in Fig.~\ref{fig:generation_types}; therefore, let $\mathcal{D}_{unknown}$ be a set of datasets with unknown deepfake generation methods:
\begin{equation}
    \label{eq:unknown}
    \small
    \mathcal{D}_{unknown}= \left \{ DFW \right\},
\end{equation}
In this work, we use the above mentioned known deepfake datasets to emulate the in-domain and out-of-domain attackers and unknown datasets to emulate the open-domain attacker models.


\begin{figure*}[t]
    \centering
    \includegraphics[width=1\linewidth]{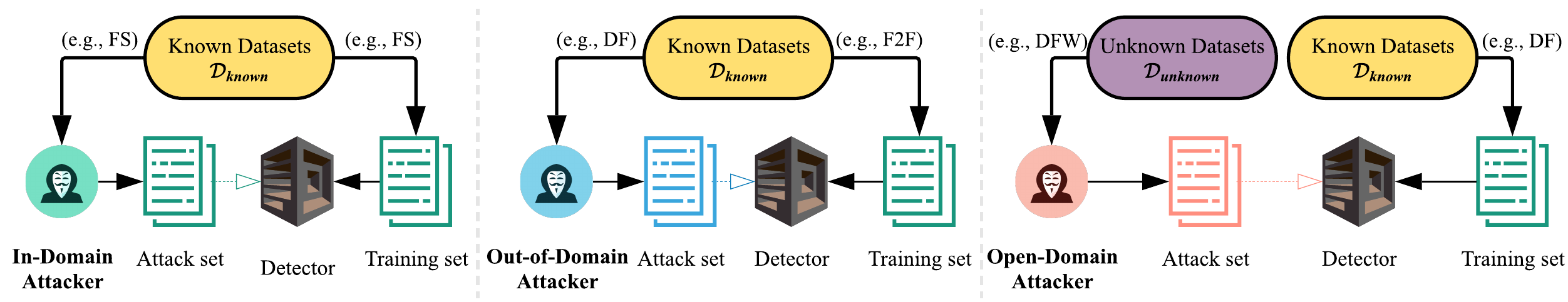}
    \caption{A visual representation of the threat model. We consider three types of attacks, each from an in-domain attacker (left), an out-of-domain attacker (middle) and an open-domain attacker (right).}
    \label{fig:Threat_Model}
\end{figure*}

\subsection{Deepfake Detection Methods} 
The detection of abnormal eye blinking~\cite{eyeblink} has shown to be effective for the identification of inconsistencies in manipulated videos or images. Furthermore, image splice detection methods~\cite{Splice1,FT36,FT6} aim to exploit the deviation resulting from splicing near the boundaries of manipulated regions in an image. Although inconsistencies in images generated by existing deepfake generation methods can be detected, new and more advanced generation methods are researched and developed every year.
Deep learning-based approaches in a supervised environment have shown high detection accuracy. Specifically, Convolutional Neural Network (CNN)-based approaches concentrated on automatically learning hierarchical representations from RGB color images input~\cite{FT27,FT33} or utilizing manipulation detection features~\cite{FT7}, and using hand-crafted features~\cite{FT11}. Tariq \textit{et al.}~\cite{Shahroz1,Shahroz2} introduced ShallowNet, a fast learning and effective CNN-based network for detecting GAN-generated images with high accuracy even at low resolution (64$\times$64). Furthermore, Zhou \textit{et al.}~\cite{FT44} applied a two-stream Faster R-CNN network, which can capture high and low-level image details.
R\"ossler \textit{et al.}~\cite{FaceForensics++} achieved a significantly improved performance on compressed images, which is essential for detecting deepfakes on social networking sites, such as Instagram, Facebook, and Twitter. Mirsky and Lee~\cite{Mirsky_WenkeLee_Survery} provided a survey of different deepfake detection methods. Most of the aforementioned approaches concentrate on detecting facial manipulations in a single video frame. However, as shown in Fig.~\ref{fig:inconsistency}, it is crucial to analyze the temporal information between consecutive frames in deepfake videos. In our approach, we use multiple consecutive frames to utilize this spatio-temporal information for an improved detection of deepfakes.

\noindent
\textbf{Attack Detection with Consecutive Frames. }
Sabir \textit{et al.}~\cite{DFD2} proposed a detection method that utilizes both CNN and Recurrent Neural Network (RNN) to capture the temporal information presented in consecutive deepfake video frames. Also, G\"uera \textit{et al.}~\cite{FT_SQ2} adopted a similar approach, extracting features from up to 80 consecutive frames using CNN layers and feeding them to RNN layers to build a temporal information-aware deepfake detection model. Both methods~\cite{DFD2,FT_SQ2} extract features from CNN and pass it to RNN layers. 
On the other hand, we build our CLRNet model using Convolutional LSTM cells, which are more inclusive, as they can capture the spatio-temporal information and extract features straight from an input image sequence, eliminating the need for passing through two different networks (CNN and RNN). Moreover, these approaches yielded poor results when evaluated on unseen deepfake generation methods. Thus, we explored transfer learning and model generalization to address this challenge.

\noindent
\textbf{Detection Generalization via Transfer Learning. }
A variety of deepfake video generation techniques are constantly being developed and more sophisticated deepfake videos will arise in the future. However, collecting and producing a significant amount of new deepfake samples would be impractical. To address this issue, few-shot transfer learning (TL) can be used for the detection of deepfakes created using different methods. That is, what has been learned in one domain (e.g., FaceSwap) can be used to enhance the generalizability in another domain (e.g., Face2Face).
Kodirov \textit{et al.}~\cite{FT25} investigated the use of an autoencoder as a method for learning better semantic representation with zero-shot or few-shot learning. Other approaches~\cite{FT8,FT17} focused on learning a latent embedding in the original domain from which the target domain's feature space is extracted. Moreover, generative adversarial networks (GANs) have also been recently used to bridge the gap between two different, but comparable domains~\cite{FT19,FT40}. In this work, we used transfer learning to enhance generalizability. Additionally, we further evaluate and compare our model with unseen real-world DFW videos, generated by unknown methods.

%% file: 3_1_AttackModel.tex
\section{Threat Models}
\label{sec:attack}

For deepfake detection and defense, we consider the following three types of deepfake generation attacker models: 1) in-domain deepfake attacker, 2) out-of-domain deepfake attacker, and 3) open-domain deepfake attacker, as shown in Fig.~\ref{fig:Threat_Model}. 
First of all, let us define in-domain, out-of-domain, and open-domain. Assuming we have three deepfake datasets, $A$, $B$, and $C$, where $A$ and $B$ are generated by known methods, such that $A\subset\ B$, $A\ne B$ (i.e., $A\subsetneq B$), while $C$ is generated by a method completely unknown to the detector. Consider the following scenario where we train a detection model using only the dataset $A$. After training, we can test (attack) the detector with the three types of datasets. If we use the dataset $A$ as input, we use in-domain data; if we use the dataset $B$ as input, we use out-of-domain data; lastly, if we use the dataset $C$, we use the open-domain data, since it is completely new.
To describe the aforementioned cases, we define the \textit{attack dataset} ($\mathcal{D}_A$) as the deepfake dataset used by the attacker, such that $\mathcal{D}_A\subset\mathcal{D}_{known}\cup\mathcal{D}_{unknown}$. Similarly, we define the \textit{training dataset} ($\mathcal{D}_T$) as the deepfake dataset to train the detector, such that $\mathcal{D}_T\subset \mathcal{D}_{known}$. 
Table~\ref{tab:Attacks} presents the three attacker models using $\mathcal{D}_A$ and $\mathcal{D}_T$, with examples. We provide more details on each attack .

\begin{table}
\centering
\caption{Attacker type and deepfake attack generation conditions for attack dataset $(\mathcal{D}_A)$ and training dataset $(\mathcal{D}_T)$.}
\label{tab:Attacks}
\resizebox{\linewidth}{!}{%
\begin{tabular}{l|l|l|l} 
\toprule
\multicolumn{1}{c|}{\begin{tabular}[c]{@{}c@{}} \textbf{Attacker}\\\textbf{Type} \end{tabular}} & \multicolumn{1}{c|}{\begin{tabular}[c]{@{}c@{}}\textbf{Attack}\\\textbf{Generation}\\\textbf{Conditions} \end{tabular}} & \multicolumn{1}{c|}{\begin{tabular}[c]{@{}c@{}}\textbf{Examples of }\\\textbf{Attack Dataset}\\\textbf{} $(\mathcal{D}_A)$ \end{tabular}} & \multicolumn{1}{c}{\begin{tabular}[c]{@{}c@{}}\textbf{Examples of}\\\textbf{Training Dataset}\\\textbf{} $(\mathcal{D}_T)$ \end{tabular}} \\ 
\hline
\begin{tabular}[c]{@{}l@{}}\textbf{In-domain}\\\textbf{Attacker} \end{tabular} & \begin{tabular}[c]{@{}l@{}}$\mathcal{D}_A\subset\mathcal{D}_{known}$\\$\mathcal{D}_T\subset\mathcal{D}_{known}$\\$\mathcal{D}_A=\mathcal{D}_T$\\ \end{tabular} & \begin{tabular}[c]{@{}l@{}}$\mathcal{D}_A=\{DF\}$\\$\mathcal{D}_A=\{FS\}$\\$\mathcal{D}_A=\{NT\}$ \end{tabular} & \begin{tabular}[c]{@{}l@{}}$\mathcal{D}_T=\{DF\}$\\$\mathcal{D}_T=\{FS\}$\\$\mathcal{D}_T=\{NT\}$ \end{tabular} \\ 
\hline
\begin{tabular}[c]{@{}l@{}}\textbf{Out-of-domain}\\\textbf{Attacker} \end{tabular} & \begin{tabular}[c]{@{}l@{}}$\mathcal{D}_A\subset\mathcal{D}_{known}$\\$\mathcal{D}_T\subset\mathcal{D}_{known}$\\$\mathcal{D}_T\subsetneq\mathcal{D}_A$ \end{tabular} & \begin{tabular}[c]{@{}l@{}}$\mathcal{D}_A=\{F2F\}$\\$\mathcal{D}_A=\{F2F, NT,$\\\qquad\quad$DFD\}$ \end{tabular} & \begin{tabular}[c]{@{}l@{}}$\mathcal{D}_T=\{DF\}$\\$\mathcal{D}_T=\{DF, FS\}$\\$\mathcal{D}_T=\{DF, FS\}$ \end{tabular} \\ 
\hline
\begin{tabular}[c]{@{}l@{}}\textbf{Open-domain}\\\textbf{Attacker} \end{tabular} & \begin{tabular}[c]{@{}l@{}}$\mathcal{D}_A\subset\mathcal{D}_{unknown}$\\$\mathcal{D}_T\subset\mathcal{D}_{known}$ \end{tabular} & \begin{tabular}[c]{@{}l@{}}$\mathcal{D}_A=\{DFW\}$\\ \end{tabular} & \begin{tabular}[c]{@{}l@{}}$\mathcal{D}_T=\{DF, FS\}$\\$\mathcal{D}_T=\{FS, DF,$\\$ DFD, NT, F2F\}$\\ \end{tabular} \\
\bottomrule
\end{tabular}%
}
\end{table}

\noindent
\textbf{In-Domain Deepfake Attacker. }
In this scenario, $\mathcal{D}_A$ and $\mathcal{D}_T$ are generated by the same method. Hence, the attacker can supply test inputs from the same deepfake dataset as the training dataset. Note that the training and attack sets have different test samples. As shown in Table~\ref{tab:Attacks}, $\mathcal{D}_A\subset\mathcal{D}_{known}$, $\mathcal{D}_T\subset\mathcal{D}_{known}$, and $\mathcal{D}_A=\mathcal{D}_T$. The 2\textsuperscript{nd} and 3\textsuperscript{rd} column of Table~\ref{tab:Attacks} shows some examples of $\mathcal{D}_A$ and $\mathcal{D}_T$ (e.g., $\mathcal{D}_A= \mathcal{D}_T= \{DF\}$).
This in-domain attack is the most widely used attack (generation) vs. detection scenario, but it is a weak attack, since $\mathcal{D}_A$ and $\mathcal{D}_T$ are the same; therefore, the deep learning-based detector performance will generally be very high, as shown in Table~\ref{tab:In-Domain_Attack}, when a large amount of the dataset is available. This attack is the first method one can use to train the detector model. Moreover, it can be used to evaluate the best performance of a detector on an individual dataset, as shown in other studies~\cite{FaceForensics++, afchar2018mesonet}.

 

\noindent
\textbf{Out-Of-Domain Deepfake Attacker. }
Considering the attacker's goal is to evade the target detector or degrading its performance, the attacker can choose those \textit{attack datasets} ($\mathcal{D}_A$) that are absent in the \textit{training datasets} ($\mathcal{D}_T$) 
as shown in Table~\ref{tab:Attacks}, i.e., $\mathcal{D}_A\subset\mathcal{D}_{known}$, $\mathcal{D}_T\subset\mathcal{D}_{known}$, and 
$\mathcal{D}_T\subsetneq\mathcal{D}_A$.
For example, when the detector is trained only on one dataset (e.g., $\mathcal{D}_T= \{DF\}$), the attacker can perform a single out-of-domain attack by choosing a different dataset 
to evade the detector (e.g., $\mathcal{D}_A= \{F2F\}$). Similarly, when the detector is trained on multiple datasets (e.g., $\mathcal{D}_T= \{DF, FS\}$), the attacker can perform a multi-out-of-domain attack by providing different examples from those in the \textit{attack dataset} to evade the detector (e.g., $\mathcal{D}_A= \{DF, NT\}$ or $\mathcal{D}_A= \{F2F, NT, DFD\}$). This out-of-domain attack is stronger than single-domain attack, as $\mathcal{D}_A$ and $\mathcal{D}_T$ are different. It is more challenging for the detector to identify this attack, as the attacker can trivially add new datasets to an attack to degrade the performance. However, adding more datasets to the detector would incur more time and the detector would particularly perform poorly on the dataset it was not trained on. We evaluate the performance of detectors against this attack.


\noindent
\textbf{Open-Domain Deepfake Attacker. } 
We can consider the most powerful attacker, which uses a completely unknown dataset \allowbreak $\mathcal{D}_{unknown}$ to evade the detector. In an open-domain attack, the attacker can try and test an unknown dataset $\mathcal{D}_A\subset\mathcal{D}_{unknown}$ on the detector only trained using known datasets $\mathcal{D}_T\subset\mathcal{D}_{known}$, where $\mathcal{D}_{known}$ is not a well-known benchmark dataset. To emulate the unknown dataset, we collected a deepfake dataset in the wild from the Internet (e.g., $\mathcal{D}_A=\{DFW\}$). In such case, we do not have precise knowledge on how those DFW videos are created. It is possible that some videos in DFW are generated by $\mathcal{D}_{known}$, but we found that many of them are different, as they yield low detection performance as shown in Table~\ref{tab:OOD_F2F} and~\ref{tab:open-domain_defense}.   
For example, an open-domain attack can be performed by using examples from an unknown dataset $\mathcal{D}_A\subset\mathcal{D}_{unknown}$ to evade the detector (e.g., $\mathcal{D}_A= \{DFW\}$), when the detector is trained on multiple known datasets $\mathcal{D}_T\subset\mathcal{D}_{known}$ (e.g., $\mathcal{D}_T= \{DF, FS, DFD\}$).
In addition, $\mathcal{D}_{unknown}$ can be used to test a model's generalizability. Since unknown deepfake generation methods will inevitably emerge in the future, it is important to evaluate the performance against real-world deepfake datasets in addition to standard benchmark datasets.

%% file: 4_approach.tex
\section{Our Method for Generalization}
\label{sec:approach}
In this section, we provide our intuition for exploring temporal information for deepfake detection, and describe the detailed architecture of CLRNet as well as defense strategies.

\begin{figure}[t!]
    \centering
    \includegraphics[width=1\columnwidth]{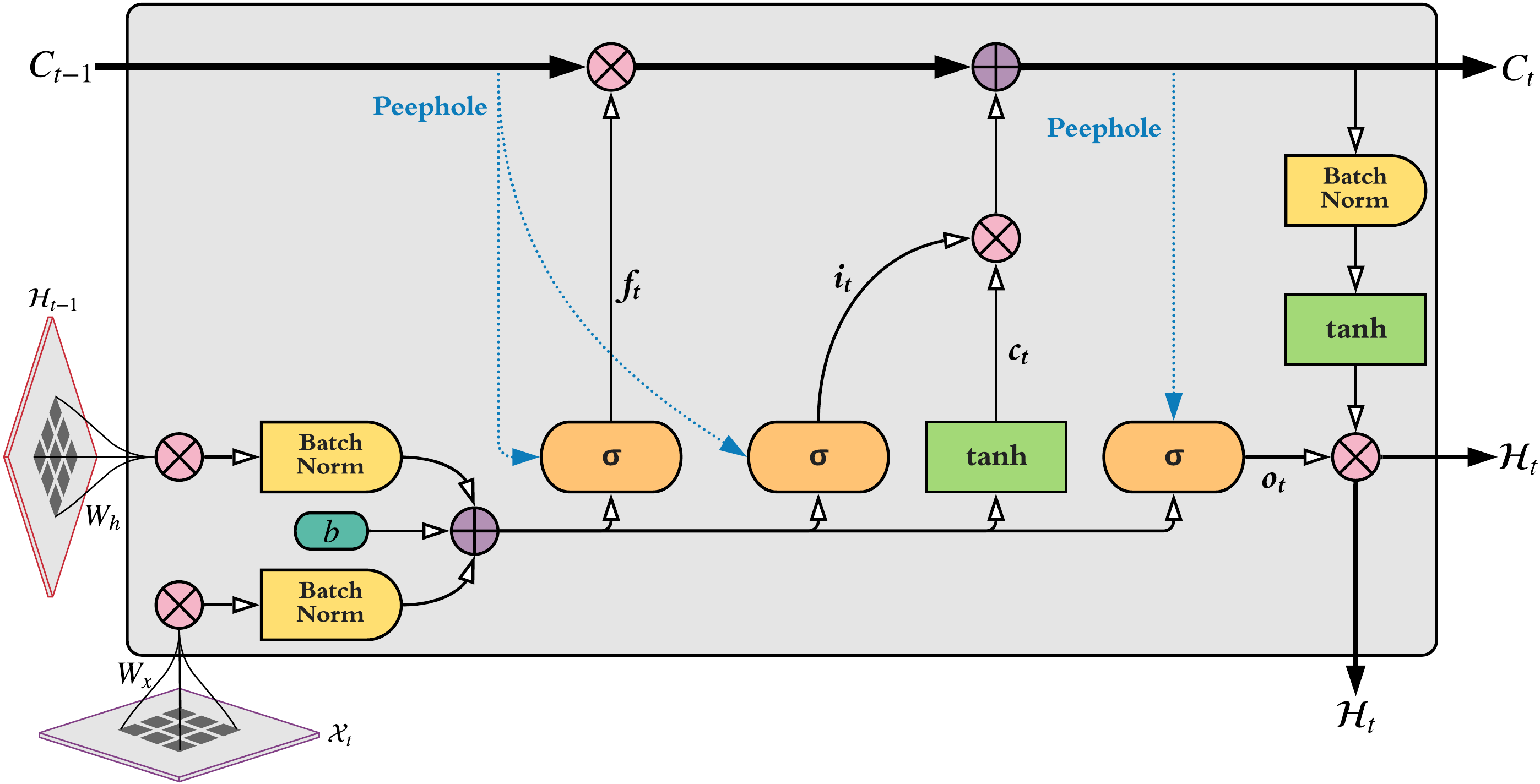}
    \caption{\emph{Illustration of Convolutional LSTM cell}: 
    Here, $\mathcal{X}_t$, $C_t$, and $\mathcal{H}_t$ is input, cell output, and hidden state. The gates are denoted by $i_t$, $f_t$, and $o_t$.}
    \label{fig:convlstm_cell}
\end{figure}

\begin{figure*}[t!]
    \centering
    \includegraphics[width=1\linewidth]{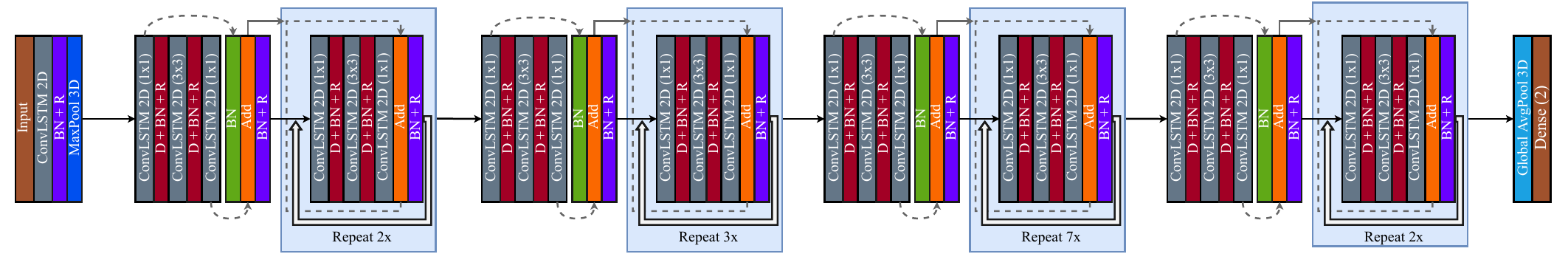}
    \caption{\emph{CLRNet Architecture}: The input to the model is a sequence of consecutive images and the output is a classification result, real or fake. We used Keras terminology to denote the layer names.}
    \label{fig:CLRNet_model}
\end{figure*}
\noindent
\textbf{Intuition. }
All of the state-of-the-art deepfake detection methods are developed on a single frame-based detection. However, they do not take into account the inter frame relationship. Therefore, we believe that these methods cannot capture the bigger picture and eventually will not be as generalizable as a method that captures both the artifacts inside a particular frame and between consecutive frames.
Based on this intuition, we performed a frame-by-frame analysis of deepfake videos and found some inconsistencies between consecutive frames in deepfake videos, which were absent in real videos. These inconsistencies include 1) a sudden change in brightness and contrast on a small region of the face, and 2) the size of some facial parts such as eyes, lips, and eyebrows changes between frames. Figure~\ref{fig:inconsistency} shows an example of such artifact from two consecutive frames of a deepfake video along with their sudden differences marked in red. These inconsistencies render the video somewhat unnatural. 
Motivated by this finding and observation, we developed a new deepfake detection method, the Convolutional LSTM Residual Network (CLRNet). Our detection method can capture 1) the differences between a real and a deepfake frame using the convolutions inside the Residual Network and 2) the inter-frame consistencies in real videos and in-consistencies in deepfake videos using the Convolutional LSTM cells. Therefore, CLRNet can learn from both the intra-frame artifacts and inter-frame artifacts, which gives a more holistic view of the data. This dual learning aspect of the CLRNet model makes it more robust and generalizable then single-frame-based methods (see Section~\ref{sec:res}).  It is a significant improvement over the single-frame-based methods which are confined to the boundaries of a particular frame. 

\subsection{CLRNet}
This section will explain the background of Convolutional LSTM Cell. Also, we will provide the architectural details of our Convolutional LSTM based Residual Network.

\noindent
\subsubsection{\textbf{Convolutional LSTM Cell.}}
Shi \textit{et al.}~\cite{ConvLSTM1} stated that the main problem with handling spatio-temporal data in Fully connected LSTM~\cite{FC-LSTM} is the use of full connections during input-to-state and state-to-state transitions, and no spatial information is encoded. In contrast, Convolutional LSTM (ConvLSTM) overcomes this problem by introducing 3D tensors whose last two dimensions are spatial (rows and columns) for all the inputs $\left (\mathcal{X}_1, \dots ,\mathcal{X}_t\right)$, outputs $\left (C_1, \dots ,C_t\right)$, hidden states $\left (\mathcal{H}_1, \dots ,\mathcal{H}_t\right)$, and gates $\left (i_t,f_t,o_t\right)$. In this paper, we follow the formulation of ConvLSTM by Shi \textit{et al.}~\cite{ConvLSTM1} (\textbf{Note:} \textit{ConvLSTM cell is different from LSTM Cell and CNN + LSTM models}). 
A visual representation of our ConvLSTM cell, based on Xavier~\cite{Convlstmcell} and implementation of Keras, is shown in Fig.~\ref{fig:convlstm_cell}.

\noindent
\subsubsection{\textbf{Convolutional LSTM Residual Network.}}
To capture the inconsistencies between consecutive video frames and artifacts within frames, we need a model that can capture temporal information like an LSTM and spatial information like a CNN. One option is to use a stacked LSTM + CNN network. However, based on our preliminary experiments, these models were unstable and do not work well with transfer learning. Therefore, we opted for another solution: Convolutional LSTM cells that serve the same purpose but have shown to be more stable~\cite{tariq2019convlstm}. We designed our model from scratch by stacking Convolutional LSTM cells and found promising results in our preliminary experiments.
Furthermore, as stated by Shi\textit{et al.}~\cite{ConvLSTM1}, Stacking ConvLSTM layers provide a robust representational power to the model. However, by stacking multiple ConvLSTM layers together, we faced the vanishing gradient problem. Residuals connections, which are first introduced by He \textit{et al}.~\cite{resnet}, are well known to solve the vanishing gradient problem. Therefore to tackle this problem, we introduced residual blocks with skip connections, as shown in Fig.~\ref{fig:CLRNet_model}.  Our model's input elements are 3D tensors, which preserve the entire spatial information for consecutive frames. We named our model as Convolutional LSTM based Residual Network (CLRNet).

\subsection{Training Strategies}


In order to defend against the in-domain, out-of-domain, and open-domain attacks, we developed three fundamental training strategies. Figure~\ref{fig:pipeline} presents an overview of these different training  methods. 

\noindent
\subsubsection{\textbf{Single Domain Learning.}} When there is only one \textit{training dataset} (e.g., $\mathcal{D}_T=\{\text{FS}\}$), we use single domain learning to train the detector, as shown in Fig.~\ref{fig:pipeline} (Middle line, $\mathcal{D}_T=\{\text{FS}\}$). 
Currently, single domain learning is a widely employed approach to detect individual deepfake generation method, as it is straightforward to train. Also, single domain learning is required as a preliminary step and primitive element for merge and transfer learning.

\noindent
\subsubsection{\textbf{Merge Learning.}} The intuition behind using this method is that the single domain learning model is trained on only one dataset; therefore, it may fail to defend against the out-of-domain attack. However, if we train the detector with all the known datasets (e.g., $\mathcal{D}_T=\{\text{FS, DF, DFD, F2F, NT}\}$), we hope that the detector can be better equipped to defend against the out-of-domain attack, as it will retain a holistic view of all the known attacks during the training. To apply merge learning, it is important to merge and aggregate multiple datasets for training.
In this method, we take the union ($\cup$) of all the training datasets together to make a single combined dataset and then use this merged dataset for training the detector, as shown at the bottom green line ($\mathcal{D}_T= \{\text{FS}\cup\text{DF}\cup\text{DFD}\cup\text{F2F}\cup\text{NT}\}$) in Fig.~\ref{fig:pipeline}. Merge learning is based on a simple idea and relatively easy to implement. However, it is the most efficient way to train, as new deepfakes arise, requiring large amount of data from each deepfake generation methods for training.  
Therefore, we consider the transfer learning, which requires much smaller training samples.   

\noindent
\subsubsection{\textbf{Transfer Learning.}} 
\begin{figure*}[t]
    \centering
    \includegraphics[clip, trim=15pt 15pt 15pt 15pt,width=0.7\linewidth]{"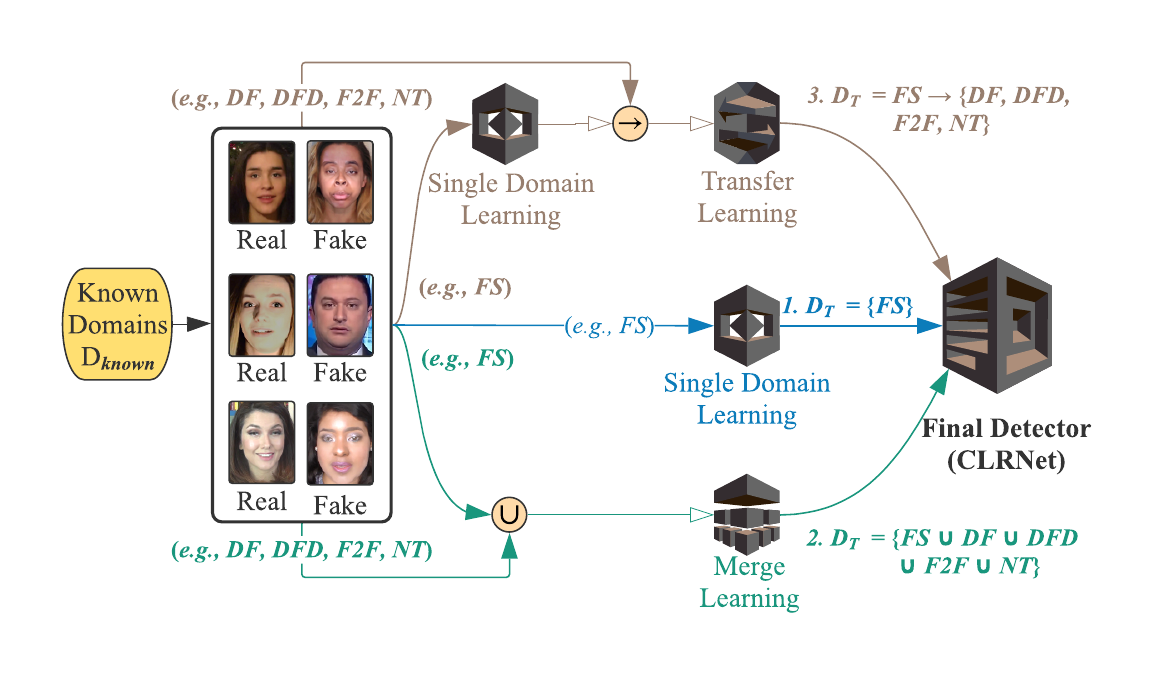}
    \caption{Overview of the three defense strategies to build CLRNet based deepfake detector against deepfake attacks.}
    \label{fig:pipeline}
\end{figure*}

The main intuition behind transfer learning is that if we increase the number of datasets in merge learning, then it may not be able to generalize well due to the excessive variety inside the single merged dataset. However, if we fully train the model on one source dataset, then the model will obtain a detailed understanding of the difference between a single type of deepfake and real video, and from there on, the model can easily extend its knowledge about different types of deepfakes by learning new domains from a small number of samples using transfer learning. 
Therefore, in this method, we first train the detector on one of the training datasets as a source (e.g., $\mathcal{D}_T=\{\text{FS}\}$ with 750 videos) using \textit{single domain learning}. Once we fully train the model on this single dataset (source), we take very few samples (e.g. 10 videos) from each of the remaining training datasets (e.g., $\{\text{DF, DFD, F2F, NT}\}$) as target datasets and use transfer learning ($\rightarrow$) to learn targets, as shown at the top brown line ($\mathcal{D}_T= \text{FS}\to\{\text{DF, DFD, F2F, NT}\}$) in Fig.~\ref{fig:pipeline}. 
Transfer learning requires freezing some layers of the deep learning model to learn efficiently. For this purpose, we set the first half of the model layers as untrainable. By doing so, we are forcing the model to learn only the small but crucial details about the new domains, which are typically present in the deeper layers of the model. Transfer learning is also known as few-shot learning, as we are only providing a few samples to the detectors to learn the new deepfakes. Therefore, we hypothesize that transfer learning is more time- and resource-efficient as it requires fewer data samples to train on new domains as compared to merge learning. We expect that it can provide at least comparable or better performance than single domain and merge learning, which we analyze in Section~\ref{sec:res}.

\subsection{Defense Strategies}
Once we have defined the different training methods, we can apply above specific training methods to detect the following in-domain, out-of-domain, and open-domain attack as follows:

\noindent
\subsubsection{\textbf{Defense against in-domain attack.}}
This attack is relatively easy to defend, as the \textit{attack dataset} ($\mathcal{D}_A$) and \textit{training dataset} ($\mathcal{D}_T$) are the same. (Note: The video samples used for $\mathcal{D}_A$ are not present in $\mathcal{D}_T$.) Therefore, we can apply a single domain learning technique to achieve high performance against this attack (see Section~\ref{sec:res} and Table~\ref{tab:In-Domain_Attack}), similarly as presented in prior work~\cite{FaceForensics++, afchar2018mesonet}.

\noindent
\subsubsection{\textbf{Defense against out-of-domain attack.}} These attacks are from $D_{known}$ and assuming that the detector is not trained on all the possible datasets in $D_{known}$. Therefore, the out-of-domain attack is more powerful attack and can decrease the performance (F1-score) of a detector trained on a single domain learning. The best strategy for the defender would be to use all the known dataset $D_{known}$ to train the detector via merge or transfer learning. For merge learning, we can combine all the datasets (e.g., $\text{FS}\cup\text{DF}\cup\text{DFD}\cup\text{F2F}\cup\text{NT}$) and train the detector. However, for transfer learning, we can randomly select one dataset (e.g., $FS$) and train on it using single domain learning and then apply the transfer learning to learn the rest of the domains (e.g., $\text{FS}\to\{\text{DF, DFD, F2F, NT}\}$). Therefore, defending an out-of-domain attack requires a systematic training of the detector with merge or transfer learning. We analyze the advantage and disadvantages of these two methods in Section~\ref{sec:res}.

\noindent
\subsubsection{\textbf{Defense against open-domain attack.}} As we know, that open-domain attack is generated using methods that we do not know. But, we can first train our detector using the same strategy as an out-of-domain attack. If $\mathcal{D}_{unknown}$ is similar to $\mathcal{D}_{known}$, then we can well defend against open-domain attack. If $\mathcal{D}_{unknown}$ is totally unknown and new, we can selectively choose the different deepfake generation methods such as Facial Reenactment and Identity Swap for training to detect a wide range of deepfake artifacts.
However, in order to find the best performance on the open-domain attack using merge or transfer learning, we performed a grid search, where our hyperparameters are the combination of datasets from the known datasets (e.g., $FS\cup DF \cup NT$ or $FS\to\{DF,NT\}$). Moreover, we strategically select deepfake datasets that can reduce the search space and expand the detection algorithm's coverage space. To do that, we divide the known deepfake datasets into two general types 1) Facial Reenactment ($\mathcal{D}_{FacialReenact.}=\{\text{F2F, NT}\}$) and 2) Identity Swap ($\mathcal{D}_{IdentitySwap}=\{\text{DF, FS, DFD}\}$), as shown in Fig.~\ref{fig:generation_types}. We believe that it is essential to have at least one dataset from $\mathcal{D}_{FacialReenact}$ and $\mathcal{D}_{IdentitySwap}$, so that the detector can train on both general types of deepfake attacks. Selecting the best dataset combination for an open-domain attack using our restricted grid search is described in Algorithm~\ref{alg:1}. The purpose of finding the best performance using the grid search is to examine how closely a randomly selected dataset combination can perform to the best combination obtained through an exhaustive search. We choose the $\mathcal{C}_{best}$ from $\mathcal{C}$, that bears the highest classification accuracy on open-domain attack with a minimum number of dataset combinations.

\begin{algorithm}[t]
\caption{Selecting the best combination for open-domain attack using Restricted Grid Search}
\label{alg:1}
\SetKwFunction{TransferLearning}{TransferLearning}
\SetKwFunction{MergeLearning}{MergeLearning}
\SetKwFunction{AND}{and}
\SetAlgoLined
\KwData{$\mathcal{D}_{FacialReenact}$, $\mathcal{D}_{IdentitySwap}$}
$\mathcal{C}=\varnothing$\\
 \For{$i \subset \mathcal{D}_{FacialReenact}$ \AND  $j \subset \mathcal{D}_{IdentitySwap}$}{
   $\mathcal{T}=$\TransferLearning{$i$,$j$} \\
   $\mathcal{M}=$\MergeLearning{$i$,$j$} \\
  \eIf{$\mathcal{T}$ performs better than $\mathcal{M}$}{
   Append: $\mathcal{C} \leftarrow \mathcal{T}$\\
   }{
   Append: $\mathcal{C} \leftarrow \mathcal{M}$\\
  }
 }
 \KwResult{$\mathcal{C}_{best}$}

\end{algorithm}

%% file: 5_0_Experiment.tex
\section{Experiment}
\label{sec:exp}
In this section, we will describe the experiment details as well as training and testing of detection models.

\noindent
\textbf{Dataset Description. }
To compare our method with different baselines, we used pristine (Real), DeepFake (DF), FaceSwap (FS), Face2Face (F2F), Neural Textures (NT), and DeepFakeDetection (DFD) datasets. The pristine videos from FaceForensics++~\cite{FaceForensics++} dataset are used as real videos. We also evaluated our method on 200 DeepFake-in-the-Wild (DFW) videos.
Table~\ref{tab:dataset_details} summarizes all the datasets used for this paper. Except DFD, we use 1,000 videos (1,000 real and 1,000 fake) in FaceForensics++ dataset.
We used the first 750 videos out of 1,000 for training, the next 125 for validation, and the remaining of the 125 for testing. The DFD dataset contains only 363 real and 3,000 fake videos but, to avoid data imbalance, we randomly selected 300 real and 300 fake videos from it (250 for training, 25 for validation, and 25 for testing). For transfer learning, we used only 10 videos to learn each target to compare the effectiveness of transfer vs. merge learning. 
On the other hand, we used all 200 DFW videos only for testing, in order to evaluate detection model performance against unknown deepfakes in real world.

\subsection{Baseline Detection Models}
We compared our CLRNet with three state-of-the-art methods. The following is a description of the baseline methods.


\noindent
\textit{\textbf{CNN+LSTM}}: G{\"u}era \textit{et al.}~\cite{FT_SQ2} deployed a CNN stacked on top of an LSTM network to detect deepfake. 
This method uses a minimum of 20 consecutive frames.

\noindent
\textit{\textbf{DBiRNN}}: Sabir\textit{ et al.}~\cite{DFD2} used DenseNet with a bidirectional RNN to achieve high accuracy on DeepFake, FaceSwap, and Face2Face datasets. Similar to our CLRNet, this work also uses five consecutive frames for the training and testing of the model.

\noindent
\textit{\textbf{ShallowNet}}: Tariq~\textit{et al.}~\cite{Shahroz2} showed that ShallowNet~\cite{Shahroz1} achieves high accuracy in detecting GAN-generated images. 

\noindent
\textit{\textbf{Xception}}: Xception Network~\cite{Xception} is considered as the state-of-the-art deep learning model for image classification task. We used the Keras implementation of Xception.
    
\noindent
\textit{\textbf{MesoNet}}: Afchar~\textit{et al.}~\cite{afchar2018mesonet} proposed MesoNet, which is a CNN-based state-of-the-art deepfake detection method. We used the MesoInception4 model, which is their best performer. We used the code provided by authors 
to implement MesoInception4.

\begin{table}[t!]
\centering
\caption{The details of datasets used for training and testing.}
\label{tab:dataset_details}
\resizebox{1\linewidth}{!}{%
\begin{tabular}{|l|c|c|c|c|} 
\hline
\multicolumn{1}{|c|}{ \textbf{Datasets} } & \begin{tabular}[c]{@{}c@{}}\textbf{Total}\\\textbf{ Videos} \end{tabular} & \begin{tabular}[c]{@{}c@{}}\textbf{Training}\\\textbf{Videos} \end{tabular} & \begin{tabular}[c]{@{}c@{}}\textbf{Transfer}\\\textbf{Learning} \end{tabular} & \begin{tabular}[c]{@{}c@{}}\textbf{Testing}\\\textbf{Videos} \end{tabular} \\ 
\hline
Pristine (Real) & 1,000 & 750 & 10 & 250 \\ 
\hline
DeepFake (DF) & 1,000 & 750 & 10 & 250 \\ 
\hline
FaceSwap (FS) & 1,000 & 750 & 10 & 250 \\ 
\hline
Face2Face (F2F) & 1,000 & 750 & 10 & 250 \\ 
\hline
Neural Textures (NR) & 1,000 & 750 & 10 & 250 \\ 
\hline
Deepfake Detection (DFD) & 300 & 250 & 10 & 50 \\ 
\hline
DeepFake-in-the-Wild (DFW) & 200 & \multicolumn{2}{c|}{\textit{DFW is only used for testing}} & 200 \\
\hline
\end{tabular}
}
\end{table}

\noindent
\subsubsection{\textbf{Implementation Settings.}}
We used TensorFlow v1.13.1 with Keras Library on Python v3.7.5 for the implementation of our CLRNet model. 
The codes for ShallowNet~\cite{Shahroz1}, DBiRNN~\cite{DFD2}, and CNN+LSTM~\cite{FT_SQ2} are not publicly available; therefore, we implemented them and tried our best to match the original paper's experimental settings. The total number of frames used for training is equal for all methods. However, ShallowNet~\cite{Shahroz2}, Xception~\cite{Xception}, and MesoNet~\cite{afchar2018mesonet} use a single frame as input (1$\times$80 = 80 frames), whereas our CLRNet and DBiRNN uses five consecutive frames as input (5$\times$16=80 frames) and CNN+LSTM uses 20 consecutive frames as input (20$\times$4=80 frames).

\noindent
\subsubsection{\textbf{Machine Configuration and Evaluation Metrics.}}
We used Intel(R) Xeon(R) Silver 4114 CPU @ 2.20 GHz with 256.0 GB RAM and NVIDIA GeForce Titan RTX. 
The models predict the probability of a single or a group of consecutive frames being real or fake. We use Precision, Recall, and F1-Score for the evaluation (\textit{Due to space limitations, we are reporting only the F1-Scores}). Sometimes, only a few frames contain fake content in a deepfake video; therefore, we evaluate all the video frames.   

\noindent
\subsubsection{\textbf{Preprocessing and Data Augmentation.}}
From each real and fake video, we extracted 16 samples such that each sample contains five consecutive frames ($16\times5=80$ images per video). We used multi-task CNN (MTCNN)~\cite{mtcnn} to detect the face landmark information inside the extracted frame. Afterward, we used this landmark information to crop the face from the image and aligned it to the center. The average image size after cropping the faces in the datasets is $240\times240$. Therefore, we resized all frames to a $240\times240$ resolution. We also applied data augmentation techniques to diversify the training data. We varied the following conditions: 1) Brightness (-30\% to 30\%), 2) Channel shift (-50 to 50), 3) Zoom (-20\% to 20\%), 4) Rotation (-30$^{\circ}$ degrees to 30$^{\circ}$), and 5) Horizontal flip (50\% probability).

\subsection{Configuring Training Models}
\noindent
\subsubsection{\textbf{Single Domain Learning Configuration.}} We choose one known dataset and use its training data $\mathcal{D}_{known}$ to train our CLRNet model. For example, if we choose to train on DeepFake (DF), we take the first 750 videos from the DF dataset and train the detection model after completing the preprocessing step.

\noindent
\subsubsection{\textbf{Merge Learning Configuration.}} 
We aggregate the training data for merge learning. For example, to perform $\text{DF}\cup\text{FS}$, we merge the training data of DF with FS ($750+750=1500$ videos) and perform the regular training operation. The merge learning operation can at least double the training time compared to single domain learning.

\noindent
\subsubsection{\textbf{Transfer Learning Configuration.}}
To perform transfer learning, we use a pre-trained source model (e.g., source = $DF$) and a small subset of 10 videos from target domain's dataset (e.g., target = $FS$). The main idea behind this method is to fully train on one of the widely available deepfake dataset and then use transfer learning to train for the other datasets with a small amount of samples. We freeze all the layers in the first half of the neural network when performing transfer learning for CLRNet as well as all other baseline methods for a comparison.

%% file: 5_results.tex
\section{Results}
\label{sec:res}


\subsection{Performance on In-Domain Attack}
We expect that the detector can perform well against in-domain attack as $\mathcal{D}_T=\mathcal{D}_A$. The performance results are presented in Table~\ref{tab:In-Domain_Attack}, where the columns 2--7 are the results of different in-domain attacks. All of the state-of-the-art baseline methods and our CLRNet model performed exceptionally well against the in-domain attack. On average, our CLRNet is the best performer (98.61\%), and MesoNet is the $2^{nd}$ best (98.38\%). ShallowNet, DBiRNN and CNN+LSTM showed instability as they performed well for some datasets ($DF$, $FS$, and $NT$) and poor for others  ($DFD$ and $F2F$). These results are consistent with previous research~\cite{FaceForensics++, afchar2018mesonet}. 

\subsection{Performance on Out-Of-Domain Attack}
\label{sec:ood_performance}
In contrast to the in-domain attack, we expected the out-of-domain to be stronger, and would decrease the detector's performance.
Table~\ref{tab:OOD_F2F} shows the performance results of all baseline methods and our CLRNet, when trained on Face2Face ($\mathcal{D}_T=\{\text{F2F}\}$) and tested with the rest of $\mathcal{D}_{known}$. The columns 2–6 in Table~\ref{tab:OOD_F2F} are the results of different out-of-domain attacks on the detector. As expected, all the methods performed poorly with an F1-score of approximately 50\%. On the other hand, CLRNet performed slightly better with an F1-score of 64.18\%. The poor performance is expected due to the mismatch in defense and attack dataset. 
We found very similar results when we used other datasets as the detector training dataset instead of $\mathcal{D}_T=\{\text{F2F}\}$. 
Therefore, next we employed two different defense strategies against the out-of-domain attack to improve the detection performance.

\noindent
\subsubsection{\textbf{Merge Learning Defense Strategy.}}The first defense uses the merge learning technique to merge all the $\mathcal{D}_{known}$ into one dataset and train the model with all the datasets. 
Table~\ref{tab:OOD_defense} presents the performance of different state-of-the-art models and CLRNet after applying merge learning against the same attack we used on the single domain learning method in Table~\ref{tab:OOD_F2F}. We can observe that performance of every model increased after applying merge learning, which is understandable, as the models have a holistic view of deepfake attacks. On average, the best performer for merge learning is CLRNet (87.58\%), and the second-best is MesoNet (84.35\%), which better characterize the individual deepfake artifacts and combine them well for detection. On the other hand, Xception, ShallowNet, DBiRNN and CNN+LSTM are ineffective, performing poorly below 73\%, as shown in Table~\ref{tab:OOD_defense}.

\noindent
\subsubsection{\textbf{Transfer Learning Defense Strategy.}} The second defense uses the transfer learning technique in which first we fully train on one source dataset from $\mathcal{D}_{known}$ (e.g., FS) and then apply transfer learning to the rest of the target $\mathcal{D}_{known}$ datasets, only using ten videos from each remaining dataset. 
Table~\ref{tab:OOD_defense} presents the performance results after applying transfer learning to different detection models against the same out-of-domain attack we used in Table~\ref{tab:OOD_F2F}. We first fully train one dataset from $\mathcal{D}_{known}$=FS (source) and transfer-learn to the rest of the $\mathcal{D}_{known}$ (target) datasets. (Note: we also tried other combinations for transfer learning such as $DF\to\{FS, DFD, F2F, NT\}$, but due to space limitation, we only present the results of $\text{FS}$ $\to$ $\{$DF, DFD, F2F, NT$\}$.) 
As shown in Table~\ref{tab:OOD_defense}, transfer learning performance results against the out-of-domain attacks are comparable or even higher than the merge learning, in spite of using small number of target training data. The best performer for transfer learning on average is CLRNet (97.57\%), and the second-best in the MesoNet (78.52\%). Similar to merge learning, Xception, ShallowNet, DBiRNN and CNN+LSTM showed poor performance (below 66\%) with transfer learning.
From Table~\ref{tab:OOD_defense}, we observed that with fewer training samples, CLRNet achieved higher performance (97.57\%) using transfer learning than the performance of CLRNet on merge learning (87.58\%), which shows that transfer learning is a more viable training strategy.
Moreover, the consistently high performance of our CLRNet model as compared to other state-of-the-art methods, especially using transfer learning, shows that it can generalize better with less amount of data than other baseline methods, as shown in last row of Table~\ref{tab:OOD_defense}.

\begin{table*}[t]
\centering
\caption{Detection performance comparison of state-of-the-art deepfake detection methods against in-domain attack.}
\label{tab:In-Domain_Attack}
\resizebox{0.7\linewidth}{!}{%
\begin{tabular}{|l|c|c|c|l|c|c|c|} 
\hline
\multirow{3}{*}{ \textbf{Method} } & \multicolumn{7}{c|}{\textbf{Datasets $\mathcal{D}_T=\mathcal{D}_A$ (F$_1$ score \%)} } \\ 
\cline{2-8}
 & \multicolumn{4}{c|}{\textbf{Identity Swap} } & \multicolumn{2}{c|}{\textbf{Facial Reenact.} } & \multirow{2}{*}{\textbf{Avg.} } \\ 
\cline{2-7}
 & \textbf{DF}  & \textbf{FS}  & \textbf{DFD}  & \multicolumn{1}{c|}{\textbf{DFDC}} & \textbf{F2F}  & \textbf{NT}  &  \\ 
\hline
CNN+LSTM & 78.51$\pm$3.5  & 77.75$\pm$1.5  & 70.31$\pm$5.3  & 86.28$\pm$1.5 & 71.87$\pm$3.1  & 90.54$\pm$1.0  & 77.80 \\ 
\hline
DBiRNN & 80.54$\pm$2.7  & 80.56$\pm$1.8  & 82.45$\pm$3.4  & 81.94$\pm$3.5 & 73.12$\pm$6.1  & 94.38$\pm$0.3  & 82.21 \\ 
\hline
ShallowNet & 88.97$\pm$2.5  & 93.33$\pm$1.3  & 73.73$\pm$4.7  & 91.75$\pm$0.4 & 75.26$\pm$5.1  & 99.45$\pm$0.1  & 87.08 \\ 
\hline
Xception & 99.00$\pm$0.2  & 99.29$\pm$0.2  & 95.53$\pm$0.4  & 96.50$\pm$0.1 & 87.62$\pm$2.4  & 99.46$\pm$0.1  & 96.18 \\ 
\hline
MesoNet & 99.01$\pm$0.1  & 99.26$\pm$0.1  & 95.37$\pm$0.1  & 95.69$\pm$0.1 & 99.01$\pm$0.1  & 99.27$\pm$0.1  & 98.38 \\ 
\hline
\textbf{CLRNet(Ours)}  & \textbf{99.20}$\pm$0.1  & \textbf{99.50}$\pm$0.2  & \textbf{96.00}$\pm$0.1  & \textbf{96.76$\pm$0.2} & \textbf{99.20}$\pm$0.1  & \textbf{99.50}$\pm$0.1  & \textbf{98.61}  \\
\hline
\end{tabular}
}
\end{table*}

\begin{table*}
\centering
\caption{The single-domain trained detector performance against out-of-domain attack and DFW, when $\mathcal{D}_T=\{\text{F2F}\}$, $\mathcal{D}_A\subset \mathcal{D}_{known}$ and $\mathcal{D}_T\subsetneq\mathcal{D}_A$.}
\label{tab:OOD_F2F}
\resizebox{0.7\linewidth}{!}{%
\begin{tabular}{|l|c|c|c|c|c|c|} 
\hline
\multirow{3}{*}{\textbf{Method}} & \multirow{2}{*}{$\mathcal{D}_T$ } & \multicolumn{4}{c|}{\textbf{Attack Datasets $\mathcal{D}_A$ (F$_1$ score \%)}} & $\mathcal{D}_{unknown}$  \\ 
\cline{3-7}
 &  & \multicolumn{3}{c|}{\textbf{Identity Swap}} & \begin{tabular}[c]{@{}c@{}} \textbf{Facial}\\ \textbf{Reenact.} \end{tabular} & \begin{tabular}[c]{@{}c@{}}\textbf{DeepFake}\\\textbf{in the Wild}\end{tabular} \\ 
\cline{2-7}
 & \textit{F2F}  & \textbf{DF}  & \textbf{FS}  & \textbf{DFD}  & \textbf{NT}  & \textbf{DFW}  \\ 
\hline
CNN+LSTM & \textit{71.87$\pm$3.1}  & 51.12$\pm$0.1  & 50.45$\pm$0.4  & 50.25$\pm$0.1  & 46.96$\pm$0.6  & 49.14$\pm$0.5  \\ 
\hline
DBiRNN & \textit{73.12$\pm$6.1}  & 53.65$\pm$0.3  & 50.12$\pm$0.3  & 50.86$\pm$0.5  & 48.45$\pm$0.3  & 49.34$\pm$0.1  \\ 
\hline
ShallowNet & \textit{75.26$\pm$5.1}  & 55.57$\pm$0.1  & 51.37$\pm$0.6  & 51.58$\pm$0.4  & 47.29$\pm$0.1  & 50.08$\pm$0.1  \\ 
\hline
Xception & \textit{87.62$\pm$2.4}  & 52.40$\pm$0.2  & 50.10$\pm$0.2  & 50.25$\pm$1.3  & 50.52$\pm$0.2  & 50.06$\pm$0.1  \\ 
\hline
MesoNet & \textit{99.01$\pm$0.1}  & 51.90$\pm$0.5  & 50.43$\pm$0.3  & 49.73$\pm$0.2  & 50.36$\pm$0.3  & 49.77$\pm$0.2  \\ 
\hline
\textbf{CLRNet(Ours)}  & \textit{\textbf{99.20$\pm$0.1}}  & \textbf{64.18}$\pm$0.1  & \textbf{52.32}$\pm$0.1  & \textbf{56.75}$\pm$0.2  & \textbf{50.60}$\pm$0.1  & \textbf{50.59}$\pm$0.1  \\
\hline
\end{tabular}%
}
\end{table*}

\begin{table*}[t!]
\centering
\caption{Performance comparison of defense strategies (merge and transfer learning) against out-of-domain attack.}
\label{tab:OOD_defense}
\resizebox{0.7\linewidth}{!}{%
\begin{tabular}{|l|c|c|c|c|c|c|} 
\hline
\multicolumn{7}{|c|}{\textbf{Merge Learning ($\text{FS}\cup\text{DF}\cup\text{DFD}\cup\text{F2F}\cup\text{NT}$)}} \\ 
\hline
\multirow{3}{*}{\textbf{Method}} & \multicolumn{6}{c|}{Attack Datasets $\mathcal{D}_A$ (F$_1$~score \%)} \\ 
\cline{2-7}
 & \multicolumn{3}{c|}{Identity Swap} & \multicolumn{2}{c|}{Facial Reenact.} & \multirow{2}{*}{\textbf{Avg.} } \\ 
\cline{2-6}
 & FS  & \text{DF}  & \text{DFD}  & \text{F2F}  & \text{NT}  &  \\ 
\hline
CNN+LSTM & 57.45$\pm$5.1 & 63.12$\pm$2.3 & 55.54$\pm$4.1 & 53.23$\pm$7.5 & 80.12$\pm$4.1 & 61.89 \\ 
\hline
DBiRNN & 59.42$\pm$2.1 & 61.75$\pm$2.1 & 59.85$\pm$5.2 & 55.42$\pm$1.3 & 82.10$\pm$3.8 & 63.71 \\ 
\hline
ShallowNet & 55.86$\pm$7.1 & 65.16$\pm$5.4 & 50.92$\pm$1.2 & 58.84$\pm$5.2 & 86.03$\pm$1.8 & 63.36 \\ 
\hline
Xception & 73.29$\pm$5.8 & 78.57$\pm$6.2 & 54.35$\pm$4.2 & 74.17$\pm$6.3 & 82.45$\pm$2.3 & 72.57 \\ 
\hline
MesoNet & 80.17$\pm$2.6  & 84.21$\pm$1.2  & 86.27$\pm$0.2  & 82.52$\pm$3.1  & 87.60$\pm$1.9  & 84.35  \\ 
\hline
\textbf{CLRNet(Ours)} & \textbf{87.20}$\pm$0.8 & \textbf{85.72}$\pm$0.1 & \textbf{86.52}$\pm$0.1 & \textbf{89.20}$\pm$0.1 & \textbf{89.28}$\pm$0.1 & \textbf{87.58} \\ 
\hhline{|=======|}
\multicolumn{7}{|c|}{\textbf{Transfer Learning ($\text{FS}\to\{\text{DF, DFD, F2F, NT}\}$)} } \\ 
\hline
\multirow{3}{*}{\textbf{Method}} & \multicolumn{6}{c|}{\text{Attack Datasets $\mathcal{D}_A$ (F$_1$~score \%)}} \\ 
\cline{2-7}
 & \multicolumn{3}{c|}{\text{Identity Swap}} & \multicolumn{2}{c|}{\text{Facial Reenact.}} & \multirow{2}{*}{\textbf{Avg.} } \\ 
\cline{2-6}
 & \text{FS} & \text{DF} & \text{DFD} & \text{F2F} & \text{NT} &  \\ 
\hline
CNN+LSTM & 70.21$\pm$1.0 & 52.75$\pm$5.1 & 53.52$\pm$4.5 & 50.73$\pm$3.1 & 63.34$\pm$4.1 & 58.11 \\ 
\hline
DBiRNN & 67.45$\pm$2.5 & 55.95$\pm$1.0 & 50.43$\pm$7.2 & 51.79$\pm$8.5 & 66.08$\pm$5.0 & 58.34 \\ 
\hline
ShallowNet & 87.51$\pm$3.7 & 50.29$\pm$1.4 & 42.38$\pm$9.9 & 53.83$\pm$7.2 & 67.24$\pm$8.2 & 60.25 \\ 
\hline
Xception & 73.90$\pm$7.6 & 65.04$\pm$5.7 & 57.60$\pm$8.1 & 55.43$\pm$9.2 & 76.79$\pm$7.0 & 65.75 \\ 
\hline
MesoNet & 93.36$\pm$3.1 & 69.60$\pm$9.3 & 64.58$\pm$7.9 & 83.58$\pm$7.2  & 81.51$\pm$9.1  & 78.52 \\ 
\hline
\textbf{CLRNet(Ours)} & \textbf{98.70}$\pm$0.2 & \textbf{97.23}$\pm$0.3  & \textbf{97.13}$\pm$0.1  & \textbf{97.50}$\pm$0.2 & \textbf{97.30}$\pm$0.1 & \textbf{97.57}  \\
\hline
\end{tabular}
}
\end{table*}

\noindent
\subsubsection{\textbf{Merge Learning vs. Transfer Learning.}}
As we can observe from Table~\ref{tab:OOD_defense}, the performance of merge learning and transfer learning is comparable. However, let us look at some of the advantages that we can achieve over merge learning using transfer learning. The merge learning requires a lot more data for training as compared to transfer learning. For example, for merge learning DF, FS, F2F, and NT we have to use 750 videos from each domain ($750\times4=3,000$), whereas for transfer learning FS to DF, F2F, and NT altogether we have to use 750 for the source (FS) and 10 for each target domain's dataset ($750+10\times3=780$). This dataset size difference has a massive impact on the training time of these two techniques, where merge learning take 33.6 minutes per iteration on the experiments shown in Table~\ref{tab:OOD_defense} and~\ref{tab:open-domain_defense}, and transfer learning takes just 8.3 minutes per iteration. Therefore, in these experiments, training time with the merge learning technique takes four times longer than transfer learning. As a result, we expect that merge learning would require much more time and computing resources, as new deepfakes emerge in the future. Therefore, merge learning can be unrealistic and we would need a transfer learning-based strategy that can better respond to more diverse deepfakes with reasonable amount of time and computing resources.

In addition, the merge learning technique requires an equal number of samples from each dataset to train the model. Otherwise, the model will have a bias towards the majority dataset due to data imbalance. However, when a new type of attack comes, we typically have very few samples to learn about a new deepfake. Therefore, to learn a new attack using merge learning, we have to reduce the size of each dataset to match the new attack' dataset size or use other data balancing techniques. In such scenarios, transfer learning becomes a natural and viable option as it requires only a few samples from new domains to learn them. Therefore, in summary, transfer learning is a better strategy against out-of-domain attack.

\subsection{Performance on Open-Domain Attack}

The open-domain attack is very similar to the out-of-domain attack. However, the open-domain attack is from unknown datasets $\mathcal{D}_{unknown}$. Therefore, we cannot prepare our model for this attack by training on some specific $\mathcal{D}_{known}$ datasets in advance, as we did it for the out-of-domain attack. Since the open-domain attack is at least as powerful as an out-of-domain attack, a single domain-based training would bear the same poor performance as shown in the 2$^{nd}$ col. in Table~\ref{tab:open-domain_defense}.  
For this experiment, we used the detection models trained on $\mathcal{D}_{known}$ using merge learning and transfer learning. As shown in the 3rd col. in Table~\ref{tab:open-domain_defense}, the merge learning technique show reasonable performance on the open-domain attack with CLRNet being the best performer (84.95\%), while MesoNet being the 2$^{nd}$ best (78.12\%). For the transfer learning method, CLRNet being the best performer (93.86\%) and MesoNet being the 2$^{nd}$ best (75.56\%). This result shows that 1) our CLRNet model generalizes the best among all the state-of-the-art methods, and further 2) transfer learning clearly improves the overall detection model performance and generalize better.

\noindent
\subsubsection{\textbf{Comparison of Best Performer with RGS}} We also performed another analysis to find the best model for the open-domain attack using a restricted grid search (RGS). Identifying the best solution using a restricted grid search can be time-consuming and non-realistic, due to the heuristic nature of the grid search. The purpose of this experiment is to search how closely we performed to the best performance by using an arbitrarily selected combination for merge or transfer learning ($\text{NT}\to\{\text{DF, FS, DFD, F2F}\}$). As shown in Table~\ref{tab:open-domain_defense}, CLRNet is the best performer (93.86\%), and even selecting an arbitrary combination for transfer learning can lead considerably close to its best solution (95.92\%). Moreover, nearly all models performed close to their best performance with transfer learning, showing the benefit of transfer learning is not limited to just one type of source and target configurations.

%% file: 6_conclusion.tex
\section{Discussion}
\label{sec:discussion}
\noindent
\textbf{Observations. }
CLRNet is the best performer (98.61\% on avg.) for a single domain, merge learning (87.58\% on avg.), and transfer learning (97.57\% on avg.), and it does generalize well for the open-domain attack (93.86\%). Whereas the best baseline method, MesoNet, is inconsistent with its performance where it shows good performance for a single domain (98.38\%) but shows substandard performance on transfer learning (78.52\%) and open-domain attack (80.25\%), as shown in the Table~\ref{tab:In-Domain_Attack}, \ref{tab:OOD_defense} and \ref{tab:open-domain_defense}.
Xception did not perform well for the out-of-domain attack using merge (72.57\% on avg.) or transfer learning (65.75\% on average).
CLRNet consistently achieves high $F_1$ scores across all attacks and performs the best for both out-of-domain (97.57\% on avg.) and open-domain attacks (93.86\%) with transfer learning, demonstrating its generalizability. Overall, transfer learning boosts the performance of CLRNet against all attacks.
\begin{table}[t!]
\centering
\caption{Detection performance against the open-domain attack (DFW), where we used the single-domain, merge, and transfer learning defense strategies. We also report the best performance using restricted grid search.}
\label{tab:open-domain_defense}
\resizebox{1\linewidth}{!}{%
\begin{tabular}{|l|c|c|c|l|} 
\hline
\multirow{4}{*}{ \textbf{Method} } & \multicolumn{4}{c|}{\begin{tabular}[c]{@{}c@{}} \textbf{Open-Domain Attack $\mathcal{D}_A=\{DFW\}$ }\\\textbf{(F$_1$score \%)} \end{tabular}} \\ 
\cline{2-5}
 & \begin{tabular}[c]{@{}c@{}}Single Domain\\~Learning\end{tabular} & \begin{tabular}[c]{@{}c@{}}Merge\\~Learning \end{tabular} & \begin{tabular}[c]{@{}c@{}}Transfer \\Learning\end{tabular} &
 \begin{tabular}[c]{@{}c@{}}\\ \textbf{Best} \\(RGS)\end{tabular}\\ 
 
\cline{2-4}
 & \multirow{2}{*}{$\mathcal{D}_T=\{FS\}$} & $FS\cup DF \cup DFD$  & $ NT \to\{ DF, FS,$  & \multicolumn{1}{c|}{} \\
 &  & $\cup F2F \cup NT$  & $ DFD, F2F \}$  & \multicolumn{1}{c|}{} \\ 
\hline
CNN+LSTM & 45.23$\pm$1.2 & 53.75$\pm$3.1 & 55.10$\pm$4.7 & 63.64 \\ 
\hline
DBiRNN & 47.15$\pm$1.0 & 51.97$\pm$4.6 & 58.12$\pm$6.1 & 65.71 \\ 
\hline
ShallowNet & 48.36$\pm$2.1 & 57.84$\pm$7.7 & 60.35$\pm$8.3 & 70.72 \\ 
\hline
Xception & 49.12$\pm$3.2 & 71.45$\pm$5.9 & 61.41$\pm$7.2 & 72.51 \\ 
\hline
MesoNet & 49.45$\pm$1.3 & 78.12$\pm$4.8 & 75.56$\pm$6.6 & 80.25 \\ 
\hline
\textbf{CLRNet(Ours)} & \textbf{50.65}$\pm$0.9 & \textbf{84.95}$\pm$0.2  & \textbf{93.86}$\pm$0.2 & \textbf{95.92} \\
\hline
\end{tabular}%
}
\end{table}



\begin{figure*}[t!]
    \centering
    \includegraphics[width=0.7\linewidth]{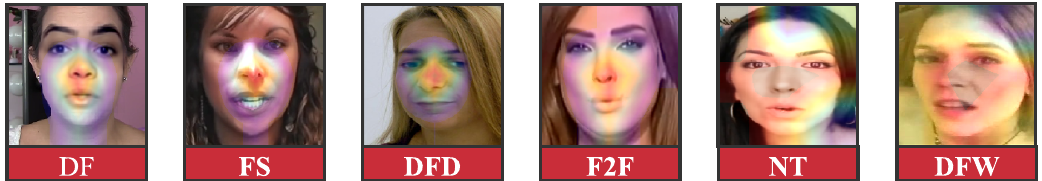}
    \caption{\emph{Class Activation Map from CLRNet}: Artifacts that distinguishes real and deepfakes are present near or around the nose region for DF, FS, F2F, and DFD. The CAM of DFW looks different for each video.}
    \label{fig:CAM}
\end{figure*}

\noindent
\textbf{Visualization using CLRNet. } 
To better characterize and analyze how CLRNet detects specific face regions of each type of deepfakes, we used the Class Activation Map (CAM)~\cite{cam}. We believe different types of deepfakes would have varying characteristics of activations. Figure~\ref{fig:CAM} presents CAM outputs of our CLRNet model for different deepfake datasets. For DF, FS, DFD, and F2F, the activations are highly concentrated on the center of the face (around the nose). However, the density and coverage vary for each dataset. For NeuralTextures (NT), the activations are randomly distributed around the face, and the area around the nose has no activations, as opposed to what we observed in other datasets, as shown in Fig.~\ref{fig:CAM}. This indicates that the DeepFake-in-the-Wild (DFW) dataset contains a mix of different deepfakes methods. We observed that some activation maps resemble NT (randomly distributed), while other approaches resemble either DF, FS, F2F, or DFD (focusing on the central part of the face). The high performance of CLRNet (93.86\%) on DFW in Table~\ref{tab:open-domain_defense}, means it is capturing all those manipulations therefore it is better generalizing and detecting different deepfake attacks with one model as compared to other methods.

\noindent
\textbf{Temporal Discontinuity. } The temporal discontinuity in most deepfakes makes them look unnatural and different from other CGI videos. However, this discontinuity can be edited out by using professional video editing tools. It is the case with high-quality DFW videos, especially the deepfake of politicians, which are very polished, but our CLRNet model is still able to detect them as deepfake using the image's spatial features. However, there are still many lifelike deepfakes that any method, including ours, can not detect.

\noindent
\textbf{Defense Aware Attacker. } In such a scenario where the attacker knows the defense method and tries to evade it, we can rely on the defense model from the open-domain attack as it is well equipped to handle the majority of the deepfakes, as shown in Table~\ref{tab:open-domain_defense}. However, further research and experiments are required in this direction.

\noindent
\textbf{Limitations and future work. } This current work does not consider different levels of video compression, where we only considered the high quality videos. We plan to experiment with different video compression levels to improve model performance and robustness. 
Also, the search space for the restricted grid search algorithm can be further improved by selecting statistically non-correlated dataset combinations to increase the detector's coverage.
In addition, detecting talking head types of deepfakes~\cite{samsungtalkinghead} are not explored in this work. It would be interesting to see how our merge and training strategy can be generalized against talking head deepfakes. Also, recently, full-body gesture-based deepfakes have emerged~\cite{FirstOrderMotion}. Therefore, we plan to detect them using our method. In this work, we also experimented using a combination of merge and transfer learning (e.g., $DF \cup NT \to FS$). However, the results were either worse or showed no improvement for all merge or transfer learning methods. Hence, future work includes the exploration of alternative training strategies that can help improve the performance.

\noindent
\textbf{Ethical Concerns. } Recently, Deepfake videos have surged, of which the vast majority uses female celebrities' face photos to develop sexually explicit videos without their knowledge or consent. These videos, rapidly spreading throughout the Internet, cause serious issues as they harass innocent people in the cyberspace. Because of the urgency of the problem and the absence of effective mechanisms to monitor and detect these videos, we have undertaken this research to investigate the real-world DeepFake-in-the-Wild (DFW) videos. For DFW video evaluation, we first consulted with the Institutional Review Board (IRB) in our institution, and they confirmed that approval was not required because the videos are already available on the Internet. In addition, we focus on obtaining freely available high quality 200 DFW videos from the Internet with celebrities and famous people, and all the researchers were informed about the detailed research protocol. Furthermore, we only used the face cropped deepfake videos for evaluation purposes. We tested 150,000 frames from 200 DFW videos of 50 celebrities and public figures.
In particular,  legal, ethical, and privacy-related issues on deepfakes are rapidly developing topics even in conducting research. Upon further consultation with a law school professor at our institution, we learned that it could be illegal to create and distribute deepfakes with malicious intents in many countries.
We did not use any child or minor deepfakes. Moreover, many news media have already presented celebrity images from deepfake videos in their articles, e.g., Daisy Ridley and Emma Watson, in BBC article~\cite{EmmawatsonDaisy}. Hence, we show that our use of images from DFW videos does not pose any legal issues.
As final verification, we reviewed the recently proposed deepfake-related laws in different countries (US, Europe, and South Korea) and found that we did not violate any laws throughout this research. Concerning privacy and ethics, we agree that some celebrities may feel offended, though some celebrities seem not to be bothered~\cite{Scarlett}. Yet, we believe it is of paramount importance and accountability to develop a technology that can stop the malicious use of deepfakes, which can go beyond celebrities and target the public in the long run. With that being said, we focus on detecting cross-domain and real-world deepfakes, of which 96\% are pornography~\cite{news5}, in addition to well-known benchmark datasets.
As researchers, we should strive to minimize any potential damage concerning ethics and privacy through our work. To reduce privacy risks, 
we will not distribute any content from DFW.


\section{Conclusion}
\label{sec:conclusion}

We introduced CLRNet, which has been successfully applied to detect various deepfakes. Instead of using a single frame, our model uses a sequence of consecutive frames from the video as an input, enabling it to capture and incorporate temporal information, and detect artifacts present within a frame and between consecutive frames. Through extensive experiments with merge and transfer learning, we were able to demonstrate the superiority of our method compared to state-of-the-art baseline approaches in terms of detection accuracy. Further, we present an effective and practical transfer learning strategy to detect multiple deepfakes simultaneously, without compromising the performance of individual deepfake detection. In summary, we addressed the shortcomings of existing methods by devising different defense strategies that show great promise in realizing more generalized deepfake attack detection models with high accuracy. Future work will explore the detection of talking head and full-body deepfake attacks. We will continue to challenge and improve existing deepfake attack detection methods to more generalizable and universal deepfake detection methods.